\documentclass[10pt,twocolumn,letterpaper]{article}

\usepackage{cvpr}              

\usepackage{graphicx}
\usepackage{amsmath}
\usepackage{amssymb}
\usepackage{booktabs}
\usepackage{multirow}
\usepackage{pifont}
\usepackage{makecell}
\usepackage[T1]{fontenc}
\usepackage{xr-hyper}

\usepackage[pagebackref,breaklinks,colorlinks]{hyperref}

\usepackage[capitalize]{cleveref}
\crefname{section}{Sec.}{Secs.}
\Crefname{section}{Section}{Sections}
\Crefname{table}{Table}{Tables}
\crefname{table}{Tab.}{Tabs.}

\def\cvprPaperID{11152} 
\def\confName{CVPR}
\def\confYear{2023}

\begin{document}

\title{Frame-Event Alignment and Fusion Network for High Frame Rate Tracking}

\author{Jiqing Zhang$^{1}$, Yuanchen Wang$^{1}$, Wenxi Liu$^2$,  Meng Li$^3$, Jinpeng Bai$^{1}$, Baocai Yin$^{1}$, Xin Yang$^{1,\star}$\\
$^1$Dalian University of Technology, $^2$Fuzhou University, $^3$HiSilicon(Shanghai) Technologies Co.,Ltd \\
}


\maketitle
\renewcommand{\thefootnote}{}
\footnote{\textsuperscript{$\star$} Xin Yang (xinyang@dlut.edu.cn) is the corresponding author.}


\begin{abstract}
Most existing RGB-based trackers target low frame rate benchmarks of around 30 frames per second.
This setting restricts the tracker's functionality in the real world, especially for fast motion. 
Event-based cameras as bioinspired sensors provide considerable potential for high frame rate tracking due to their high temporal resolution. However, event-based cameras cannot offer fine-grained texture information like conventional cameras.
This unique complementarity motivates us to combine conventional frames and events for high frame rate object tracking under various challenging conditions. In this paper, we propose an end-to-end network consisting of  multi-modality alignment and fusion modules to effectively combine meaningful information from both modalities at different measurement rates.
The alignment module  is responsible for cross-style and cross-frame-rate alignment between frame and event modalities
under the guidance of the moving cues furnished by events.
While the fusion module is accountable for emphasizing valuable features and suppressing noise information by  the  mutual complement between the two modalities.
Extensive experiments show that the proposed approach outperforms state-of-the-art trackers by  a significant margin in high frame rate tracking.  With the FE240hz dataset, our approach achieves high frame rate tracking up to 240Hz.

\end{abstract}


\section{Introduction}

 \def\wsr{1.0\linewidth}
\def\wpr{0.5\linewidth}
\begin{figure}[tbp]
	\centering
	\begin{tabular}{c}
	\includegraphics[width=\wsr]{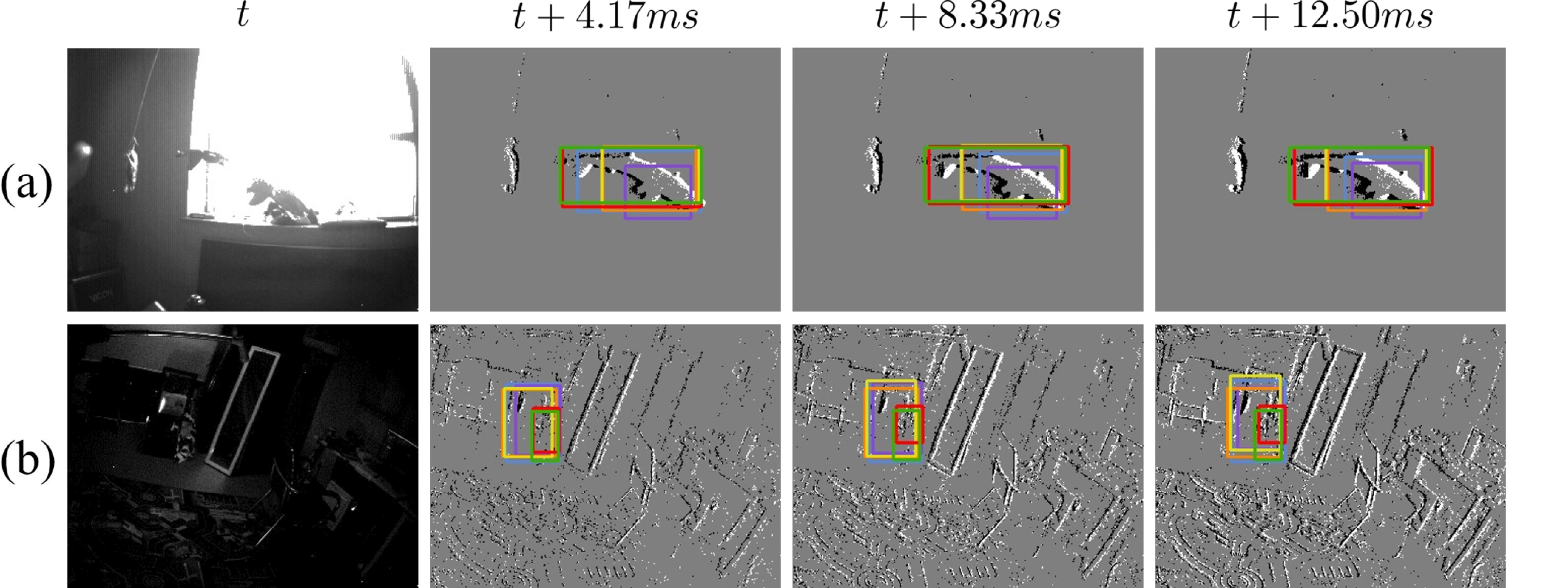}  \\ 
    \includegraphics[width=0.93\linewidth]{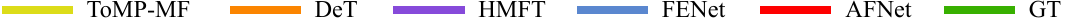}  \\ 
	\end{tabular}
	\caption{ A comparison of our  AFNet with SOTA trackers. All competing trackers locate the target  at time $t + \Delta t$ with  conventional frames at time $t$ and aggregated events  at time $t + \Delta t$ as inputs. Our method achieves high frame rate tracking up to 240Hz on the FE240hz dataset. The two examples also show the complementary benefits of both modalities. (a) The event modality does not suffer from HDR, but the frame  does; (b) The frame modality provides rich texture information, while the events are sparse. } 
	\label{fig:teaser}
\end{figure}

Visual object tracking is a fundamental task in computer vision, and deep learning-based methods~\cite{danelljan2019atom, danelljan2020probabilistic, yan2021learning, chen2021transformer, gao2022aiatrack, mayer2022transforming} have dominated this field. Limited by the conventional sensor, most existing approaches are designed and evaluated on benchmarks~\cite{7001050,8265440,mueller2016benchmark,fan2019lasot } with a low frame rate of approximately 30 frames per second (FPS). However, the 
value of a higher frame rate tracking in the real world has been proved~\cite{kowdle2018need,  handa2012real, kim2016real, kiani2017need}. For example, the shuttlecock can reach speeds of up to $493km/h$, and analyzing its position is essential for athletes to learn how to improve their skills~\cite{shum2005tracking}. 
Utilizing professional high-speed cameras is one strategy for high frame rate tracking, but these cameras are inaccessible to casual users. Consumer devices with cameras, such as smartphones, have made attempts  to  integrate sensors with similar functionalities into their systems. However,  these sensors still suffer from large memory requirements and high power consumption~\cite{tulyakov2021time}.

As bio-inspired sensors, event-based cameras  measure light intensity
changes and output asynchronous events to represent visual information. Compared with conventional frame-based sensors, event-based cameras offer a high measurement rate (up to 1MHz), high dynamic range (140 dB vs. 60 dB), low power consumption, and high pixel bandwidth (on the order of kHz)~\cite{gallego2020event}. These unique properties offer great potential for  higher frame rate tracking in challenging conditions. Nevertheless, event-based cameras cannot measure fine-grained texture information like conventional cameras, thus inhibiting tracking performance.
Therefore, in this paper, we exploit to integrate the valuable information from event-based modality with that of frame-based modality for  high frame rate single object tracking under various challenging conditions.

 
To attain our objective, two challenges require to be addressed:
(i) The measurement rate  of event-based cameras is much higher than that of conventional cameras. Hence for high frame rate tracking, low-frequency frames must be aligned with high-frequency events to disambiguate target locations. Although recent works~\cite{tian2020tdan, wang2019edvr, shi2021video, marin2022drhdr} have proposed various alignment strategies across multiple frames for video-related tasks, they are specifically designed for conventional frames of the same modality at different moments. Thus, applying these approaches directly to our cross-modality alignment does not offer an effective solution.
(ii) Effectively fusing complementary information between modalities and preventing interference from noise is another challenge. Recently, Zhang \textit{et al.}~\cite{zhang2021object} proposed a cross-domain attention scheme to fuse visual cues from frame  and event modalities for improving the single object
tracking performance under different degraded conditions. However, the tracking frequency is bounded  by the conventional frame rate since they ignore the rich temporal information recorded in the event modality.

To tackle the above challenges, we propose a novel end-to-end framework to effectively combine complementary information from two modalities at different measurement rates  for high frame rate tracking, dubbed AFNet, which consists of two key components for alignment and fusion, respectively.  Specifically, (i) we first propose an event-guided cross-modality alignment (ECA) module to simultaneously accomplish  cross-style alignment and cross-frame-rate alignment. 
Cross-style alignment is enforced by matching feature statistics
between conventional frame modality and events augmented by a well-designed attention scheme; 
Cross-frame-rate alignment is based on deformable convolution~\cite{dai2017deformable}  to   facilitate alignment without explicit motion estimation or image warping operation by implicitly focusing on motion cues.
(ii) A cross-correlation fusion (CF) module is further presented to combine complementary information by learning a dynamic filter from one
modality that contributes to the feature expression of another modality, thereby emphasizing valuable information and suppressing interference.
Extensive experiments on  different event-based tracking datasets validate the effectiveness of the proposed approach (see Figure~\ref{fig:teaser} as
an example).

In summary, we make the following contributions:

$\bullet$ Our AFNet is, to our knowledge, the first to combine the rich textural clues of frames with the high temporal resolution offered by events for high frame rate object tracking.

$\bullet$  We design a novel event-guided alignment framework that performs cross-modality and cross-frame-rate alignment simultaneously, as well as a cross-correlation fusion architecture that complements the two modalities.

$\bullet$  Through extensive experiments, we show that the proposed approach outperforms state-of-the-art trackers in various challenging conditions.

\section{Related Work}

\subsection{Visual Object Tracking}

Visual object tracking based on the conventional frame has undergone astonishing progress in recent years, which  can be generally divided into two categories, \textit{i.e.}, correlation filter (CF) trackers~\cite{bertinetto2016staple,bolme2010visual, henriques2012exploiting, ma2015hierarchical}, and deep trackers~\cite{nam2016learning, bertinetto2016fully, li2018high, yan2022towards, zhou2022global, zhou2022global1}. CF trackers learn a filter corresponding to the object of interest in the first frame, and this filter is used to locate the target in subsequent frames. While mainstream deep trackers estimate a general similarity map by cross-correlation between template and search images.
However, limited by sensors and benchmarks, those methods are mainly applied to low frame rate (30FPS) tracking.

The high temporal resolution of event cameras allows tracking targets at a higher frame rate.
Compared with conventional frame-based tracking, a few attempts have been made at event-based tracking, which can be generally classified into cluster-based and learning-based methods. 
Litzenberger \textit{et al.}~\cite{litzenberger2006embedded} assigned each new event to a cluster based on distance criteria, which is continuously updated for locating the target. Linares \textit{et al.}~\cite{linares2015usb3} used software to initialize the size and location of clusters, then proposed an FPGA-based framework for tracking. Piatkowska \textit{et al.}~\cite{pikatkowska2012spatiotemporal} extended the  clustering method by a stochastic prediction of the objects’ states to 
locate multiple persons. However, these methods involve handcrafted strategies and only apply in simple situations.
Based on the powerful representation ability of deep learning~\cite{liu2022explore,wang2023geometrical}, Chen \textit{et al.}~\cite{chen2019asynchronous,chen2020end} designed two different event representation algorithms based on Time Surface~\cite{lagorce2016hots} for target location regression. Zhang \textit{et al.}~\cite{zhang2022spiking} combined Swin-Transformer~\cite{liu2021swin} and spiking neural network~\cite{dingbiologically,liu2021event,haiwei2023} to extract spatial and temporal features for improving event-based tracking performance. However, these event-based trackers 
often fail to locate targets accurately  when events are too sparse or insufficient.

To combine benefits from  frame and event modalities, ~\cite{zhang2021object}  employed attention schemes~\cite{mei2021camouflaged,zhang2021two,mei2022glass,qiao2022cpral,qiao2020attention} to balance the contributions of the two modalities.  This work is most closely related to ours, but it does not exploit the high measurement rate of event-based cameras to accomplish a higher frame rate tracking, thus the tracking frequency is constrained by the frame rate in the frame modality. 
In contrast,  our approach attains high frame rate tracking under various challenging conditions by  aligning and fusing frame and event modalities with different measurement rates.


\subsection{Alignment  between Multiple Frames}
Alignment across multiple frames in the same sequence is  essential to exploit the temporal information for video-related tasks, such as video super-resolution~\cite{wang2019edvr, tian2020tdan} and compressed video quality enhancement~\cite{deng2020spatio, zheng2021adaptive}. 
A line of works~\cite{sajjadi2018frame,xue2019video,tao2017detail} performs alignment by  estimating the optical  flow field between the reference and its neighbouring frames. 
In another line of works~\cite{shi2021video, tian2020tdan, deng2020spatio}, implicit motion compensation is accomplished by deformable convolution.  Deformable convolution was first proposed in ~\cite{dai2017deformable}, which improves the ability of convolutional layers to model geometric transformations by learning additional offsets.  
Although the deformable convolution has shown superiority in alignment on the conventional frame domain, aligning on the frame and event modalities brings unique challenges caused by the different styles. In this paper, we propose a novel alignment strategy to simultaneously achieve cross-modality and cross-frame-rate alignment. 
 
\section{Methodology}

\def\w{0.98\linewidth}
\def\h{2.5in}
\begin{figure*}[tbp]
	\setlength{\tabcolsep}{1.0pt}
	\centering
	\begin{tabular}{c}
		\includegraphics[width=\w]{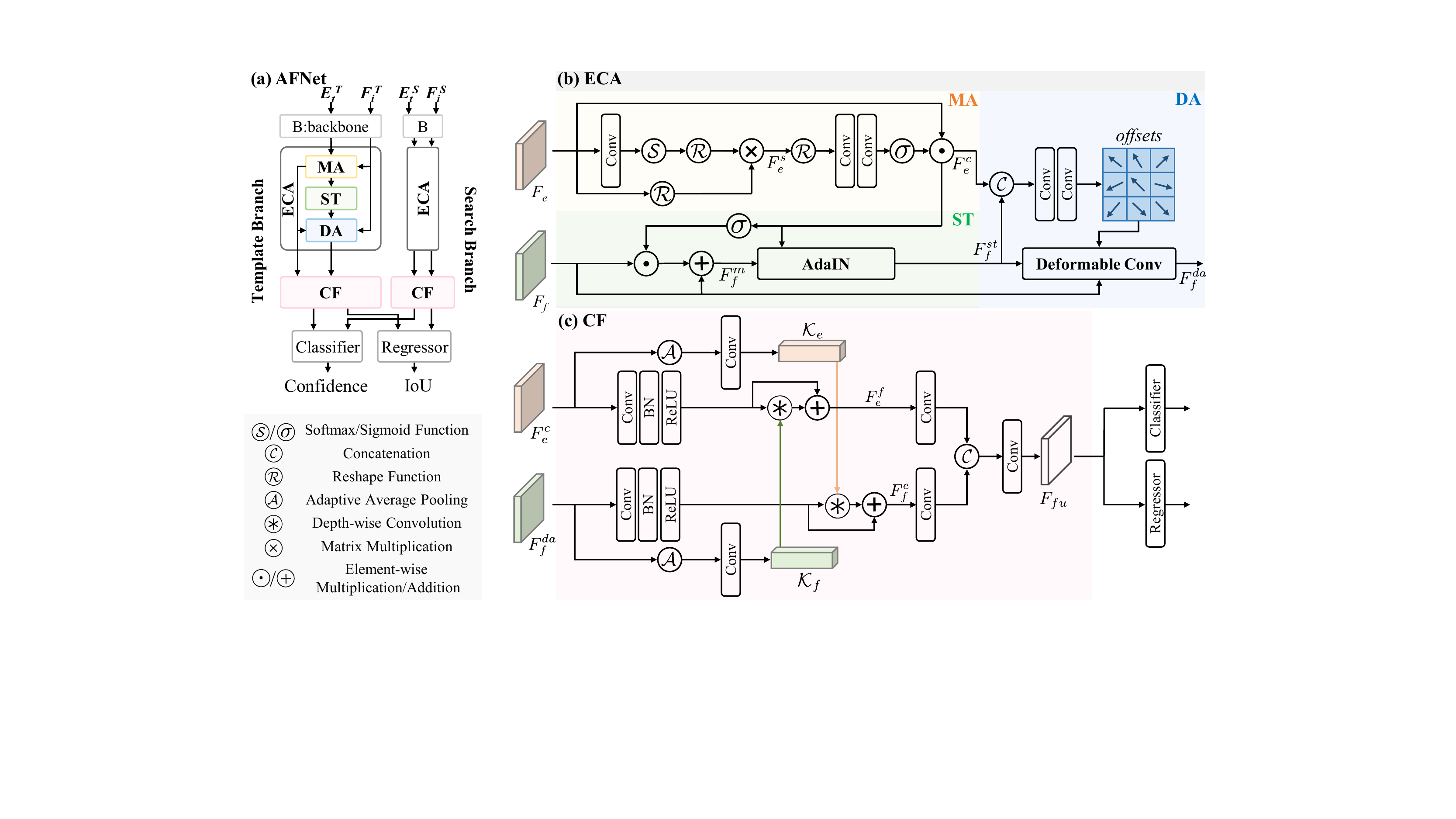} \\
 
	\end{tabular}
	\caption{(a) Overview of our AFNet;  (b) two key components in the  event-guided cross-modality alignment (ECA) module: style transformer (ST) and deformable alignment (DA); and (c) the cross-correlation fusion (CF) module.}
	\label{fig:pipeline}
	\vspace{-0.40cm}
\end{figure*}

\subsection{Events Representation}
Event-based cameras asynchronously capture log intensity change for each pixel. An event will be triggered when:
\begin{equation}
L(x, y, t)-L(x, y, t-\Delta t) \geq p C ,
\end{equation}
where  $C$ denotes a certain contrast threshold; $p$ is the polarity which means the sign of bright change, with $+$1 and $-$1 representing the positive and negative events, respectively.  $\Delta t$ is the time since the last event at location $(x, y)^\top$. 

Suppose two sequential conventional frames $F_i$ and $F_{i+1}$ are captured at times $i$ and $i+1$, respectively. $E_{i\rightarrow i+1} = \left\{\left[x_k, y_k, t_k, p_k\right]\right\}_{k=0}^{N-1}$ contains $N$ events triggered during the interval $[i, i+1]$.
Our goal is to achieve high frame rate tracking by aligning and fusing conventional frame $F_i$ and $E_{i\rightarrow t}$ at any time $t \in [i, i+1]$. 
The apart in time between dual-modality  inputs  depends on their frame rates. Specifically, $t -i = \frac{n}{\gamma_e}$, where $n$ is an integer in  $[1, \frac{\gamma_e}{\gamma_f}]$; $\gamma_e$ and $\gamma_f$ denote the frame rates of event and frame modalities, respectively. Following ~\cite{zhang2021object}, we represent events $E_{i\rightarrow t}$ as:
\begin{equation}
    g(x, y) =  \lfloor\frac{p_k\times\delta(x-x_k, y-y_k, t-t_k) + 1}{2} \times 255\rfloor ,\\ 
\end{equation}
where $g(x,y)$ denotes the pixel value of aggregated events at $(x, y)^\top$; $\delta$ is the Dirac delta function. In this way, the asynchronous event stream $E_{i\rightarrow t}$ is accumulated to a 2D event frame, denoted $E_t$.

\subsection{Network Overview}

Following DiMP~\cite{bhat2019learning}, as illustrated in Figure~\ref{fig:pipeline} (a), the overall architecture of our proposed AFNet contains three components: the feature extractor (\textit{i.e.}, backbone, ECA, and CF), the target classifier, and the bbox regressor. The feature extractors of the template branch and the search branch share the same architecture.
Each branch receives an RGB image  $F_i$ and aggregated events $E_t$  at different times as inputs, and corresponding features $F_f$ and $F_e$ can be extracted by the backbone network.  
ECA and CF are two key components of our method.  The goal of ECA is to address the misalignment between the conventional frames and aggregated event frames at different moments. While CF aims to combine the strengths of 
both modalities by complementing one modality with information from another. 
Both target classifier and bbox regressor receive the fused features from feature extractors. Given a template set of fused features and corresponding
target boxes, the model predictor generates the weights of the target classifier. 
Applying these weights to the features collected from search branch predicts the target confidence scores. The bbox regressor estimates the IoU of the groundtruth and the predicted bounding box.

 

\subsection{Event-guided Cross-modality Alignment}
The ECA module is proposed  to  align conventional frames to the reference aggregated events at the feature level.
The key to ECA is designed based on the following challenges: (i)  Cross-style alignment is a challenge. Frames and events are recorded by different sensors and thus have different styles, making alignment challenging. (ii) Cross-frame-rate alignment is another challenge. The frame rate of aggregated event frames is far higher than that of conventional images, resulting in  target location ambiguity that confuses the tracker's predictions. As shown in Figure~\ref{fig:pipeline} (b), ECA contains three modules: Motion Aware (MA), Style Transformer (ST), and Deformable Alignment (DA).

\noindent
\textbf{MA.} Since event-based cameras respond to changes in light intensity, they provide natural motion cues that can effectively facilitate multi-modality alignment.
We thus first enhance the valuable motion information of event modality by visual attention mechanisms.
As shown in Figure~\ref{fig:pipeline} (b), given event modality features $F_e \in \mathbb{R}^{C \times H \times W}$,  we design  spatial and channel attention schemes  to emphasize the meaningful moving cues while suppressing noise,
\begin{align}
    F_e^c &=  \sigma(\psi{_1}(\psi{_1}(\mathcal{R}^{(C,1,1)}(F_e^s))))F_e , \\
    F_e^s &= \mathcal{R}^{(1, C,HW)}(F_e) \times \mathcal{R}^{(1, HW,1)}(\mathcal{S}(\psi{_1}(F_e))) ,
\end{align}
where $F_e^s$ and $F_e^c$ are  event features enhanced in the spatial and channel dimensions, respectively. $\psi{_k}$ denotes the  convolution operation where kernel size is $k\times k$; $\mathcal{S}$ and $\sigma$ denote the softmax and the sigmoid function, respectively;  $\mathcal{R}(\cdot)$ is a reshape function with a target shape $(\cdot)$. 

 \noindent
\textbf{ST.} ST is responsible for combining the content of conventional frames and the style of events to meet the first challenge. Specifically, $F_e^c$ is employed to guide the frame features $F_f$  to focus on the motion cues that aid in alignment, 
\begin{align}
    F_f^m &=  \sigma(F_e^c)F_f  + F_f ,
\end{align}
where $F_f^m$ denotes frame features fused with moving information provided by events. Then, we adopt the adaptive instance normalization (AdaIN)~\cite{huang2017arbitrary} to adjust the mean and variance of the content input (\textit{i.e.}, frame features) to match those of the style input (\textit{i.e.}, event features). Formally,
\begin{equation}
\begin{aligned}
F_f^{st} &= \operatorname{AdaIN}(F_f^m, F_e^c) \\
&=\sigma(F_e^c)\left(\frac{F_f^m-\mu(F_f^m)}{\sigma(F_f^m)}\right)+\mu(F_e^c) ,
\end{aligned}
\end{equation}
where $F_f^{st}$ denotes the output of our ST module, which combines the content of frame modality and the style of event modality.
$\mu$ and $\sigma$ are the mean and standard deviation, computed independently across batch size and spatial dimensions for each feature channel.

 \noindent
\textbf{DA.} To address the second challenge, inspired by~\cite{wang2019edvr}, we propose the DA module  to adaptively align the conventional frames and aggregated events at different frame rates without explicit motion estimation and image warping operations. As shown in Figure~\ref{fig:pipeline} (b), DA first predict the offsets $\mathcal{O}$ of the convolution kernels according to  $F_e^{c}$ and  $F_f^{st}$, 
\begin{equation}
\begin{aligned}
\mathcal{O} &= \psi_3(\psi_1([F_e^c, F_f^{st}])) ,
\end{aligned}
\end{equation}
where  $[\cdot]$ denotes channel-wise concatenation. The learnable offsets will implicitly focus on motion cues and explore similar features across modalities for alignment.
With $\mathcal{O}$ and $F_f$, the aligned feature $F_f^{da}$ of the conventional frame can be computed by the deformable convolution $\mathcal{D}$~\cite{dai2017deformable},
\begin{equation}
\begin{aligned}
F_f^{da} &=\mathcal{D}(F_f, \mathcal{O})  .
\end{aligned}
\end{equation}

\subsection{Cross-correlation Fusion}
Our CF is proposed to  robustly fuse frame and event correlations by adaptively learning a dynamic filter from one modality that contributes to the feature expression of another modality.
Simply fusing frame and event modalities ignores circumstance in which one of the modalities does not provide meaningful information.
In an HDR scene, for instance, the frame modality will provide no useful information, yet the event modality still exhibits strong cues. 
Conversely, in the absence of motion, event-based cameras cannot successfully  record target-related information, while conventional frames can still deliver rich texture features.
Therefore,  we propose a cross-correlation scheme  to complement one domain with information from another domain as shown in Figure~\ref{fig:pipeline} (c).
Specifically, given the aligned frame feature $F_f^{da}$ and enhanced event feature $F_e^{c}$, the proposed CF first adaptively estimates a dynamic filter of high-level contextual information from one modality. Then, this dynamic filter serves to enhance the features of another modality. Formally, 
\begin{equation}
\begin{aligned}
F_f^{e} &= \mathcal{F} \circledast \mathcal{K}_e + \mathcal{F} ,  \\
\mathcal{F} & = \vartheta(\psi_3(F_f^{da})), \\
\mathcal{K}_e &= \psi_3(\mathcal{A}(F_e^{c})) ,
\end{aligned}
\end{equation}
where $F_f^{e}$ denotes the enhanced feature of the frame modality based on the dynamic filter $\mathcal{K}_e$ from event modality;  $\circledast$ is the depthwise convolution; $\mathcal{A}$ denotes the adaptive average pooling; $\vartheta$ is the  Batch Normalization (BN) followed by a ReLU activation function. Similarly, we can extract the complementary feature $F_e^{f}$ of the event modality    based on the dynamic filter $\mathcal{K}_f$ from 
frame modality.  Finally,  $F_f^{e}$ and $F_e^{f}$ are concatenated to build the fused feature $F_{fu}$, 
\begin{equation}
F_{fu} = \psi_1([\psi_1(F_f^{e}),\psi_1(F_e^{f})]) ,
\end{equation}
 $F_{fu}$ will be fed into the classifier and regressor to locate the target. The classifier  adopts an effective model initializer  and a steepest descent based optimizer to  predict the score map. The regressor employs the overlap maximization strategy for the task of accurate bounding box estimation.  We refer to~\cite{bhat2019learning} for details.

\subsection{Implementation Details}
We adopt the pretrained ResNet18~\cite{he2016deep}  as the backbone to extract frame and event features. Following ~\cite{bhat2019learning, zhang2021object}, the loss function is defined as: 
\begin{equation}
L = \beta L_{cls} + L_{bb} ,
\end{equation}
where $L_{cls}$ is the target classification loss which includes a hinge function to equally focus on both positive and negative samples.
$L_{bb}$ is the bounding box regressor loss which estimates MSE
between the predicted IoU and the groundtruth. $\beta$ is set to 100.

We implemented our approach in PyTorch~\cite{paszke2019pytorch} and trained our network for 100 epochs with a batch size of 32 using Adam optimizer with the default parameters. 
We set the initial learning rate of the feature extraction network,
the classifier, and  the regressor to 2e-4, 1e-3, 1e-3, respectively. The learning rate is adjusted by the  CosineAnnealingLR strategy~\cite{loshchilov2016sgdr}.
Our network is  run on a single Nvidia RTX3090 GPU with 24G memory.
 
\section{Experiments}

\subsection{Datasets}
We evaluate our AFNet on two event-frame-based datasets: FE240hz~\cite{zhang2021object} and VisEvent~\cite{wang2021viseventbenchmark}. The FE240hz dataset has annotation frequencies as high as 240 Hz and  consists of more than 143K images and corresponding recorded events. With this dataset, our method can accomplish a high frame rate tracking of 240Hz. 
Compared with FE240hz, VisEvent provides a low annotation frequency, about 25Hz. However, it contains various rigid and non-rigid targets both indoors and outdoors. 
Following~\cite{zhang2022spiking}, there are 205 sequences for training and 172 for testing.

\def\wsrpr{1.0\linewidth}
\def\hsrpr{1.2in}
\begin{figure}[t!]
	\setlength{\tabcolsep}{1.0pt}
	\centering
	\small
	\scalebox{0.95}{
	\begin{tabular}{c}
		
		\includegraphics[width=\wsrpr]{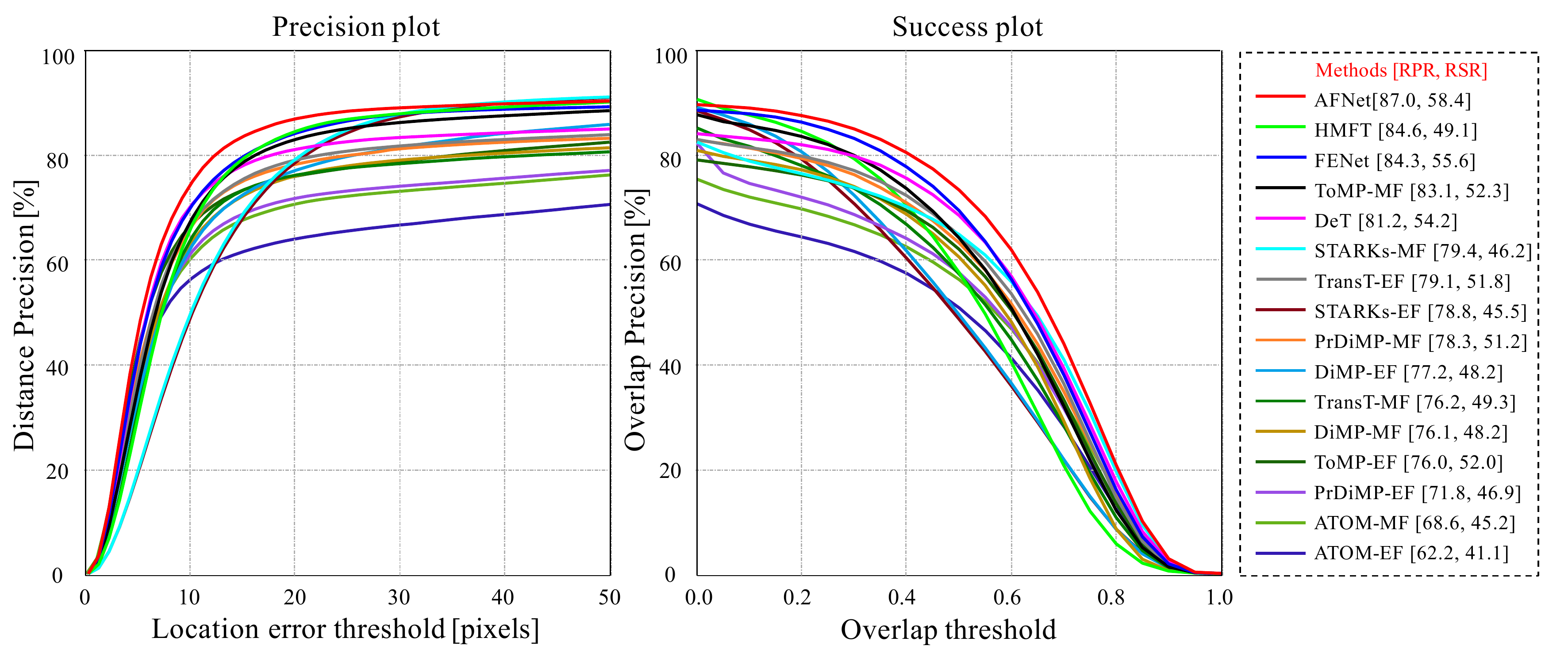}  \\
		(a) Precision  and Success  plot on FE240hz  dataset \\
		\includegraphics[width=\wsrpr]{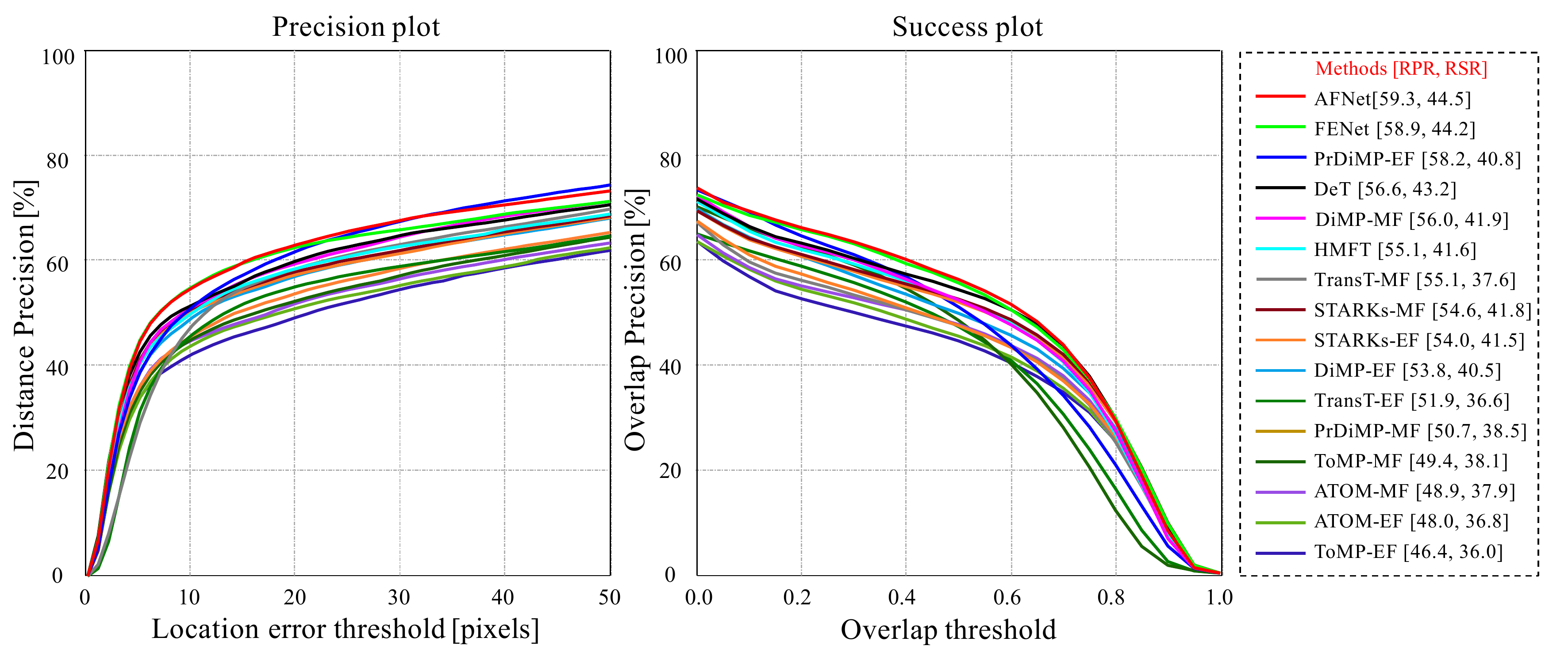}  \\ 
		(b) Precision   and Success   plot on VisEvent dataset

	\end{tabular}}
	\caption{Results on FE240hz~\cite{zhang2021object} and  VisEvent~\cite{wang2021viseventbenchmark} datasets.}
	\label{fig:PRSR}
	\vspace{-0.5cm}
\end{figure}

\setlength{\tabcolsep}{6.4pt}
\begin{table*}[t]
	\centering
	
		    \scalebox{0.95}{
	\begin{tabular}{l|c|cc|cc|cc|cc|cc|cc}
		\hline
		\hline
		  \multirow{2}{*}{Methods} &    Fusion  & \multicolumn{2}{c|}{HDR} & \multicolumn{2}{c|}{LL} & \multicolumn{2}{c|}{FM}   & \multicolumn{2}{c|}{NM} & \multicolumn{2}{c|}{SBM}     & \multicolumn{2}{c}{All} \\
		\cline{3-14} 
		 & Type & RSR  & RPR   & RSR   & RPR  & RSR  & RPR   & RSR  & RPR  & RSR  & RPR  & RSR  & RPR  \\ 
		\hline
		
		 \multirow{2}{*}{ATOM~\cite{danelljan2019atom}} &  EF & 26.6   & 42.6  &  44.6   &  67.2 &  56.4  & 83.2 &  46.7  & 78.7  &   28.9 &  41.9 & 41.1  & 62.2 \\ 
		\cline{2-14}
		 & MF &  29.1 & 48.4 &  52.7   & 78.1 &  45.4  & 68.9 &  60.8  & 92.4  & 40.1  & 60.3 &  45.2 & 68.6 \\ 
	   \hline
	   
	   \multirow{2}{*}{DiMP~\cite{bhat2019learning}}   &EF  &  38.5  & 61.0  & 58.1 & 86.8 &   53.7  & 90.4 & 50.1  & 86.5 &  47.8  & 74.8 & 48.2  & 77.2  \\ 
		\cline{2-14}
	 & MF  & 39.1   & 61.4  &  55.4  &   83.6 & 59.4  & 93.0 & 42.6   & 76.6  &  50.4  &  78.7 &  48.2 & 76.1 \\ 
	 \hline
		
		 \multirow{2}{*}{PrDiMP~\cite{danelljan2020probabilistic}}   &EF  & 22.3  &  32.7  & 64.0 & 92.2  &  53.1   & 85.0  &  56.9  & 91.8  &  35.0  & 52.9 & 46.9  & 71.8  \\ 
		\cline{2-14}
	 & MF &  39.3  & 64.1   & 63.0    &  89.3 & 60.4 &  95.7  & 55.0  & 92.2  &  47.9  & 73.8 & 51.2 & 78.3 \\ 
		 \hline
		
    \multirow{2}{*}{STARKs~\cite{yan2021learning}}   &EF  &  42.2  & 73.1  &  55.0   & 90.5 & 41.6 &  75.1  &  26.4  & 53.0  &  51.9  & 84.5 &  45.5 & 78.8 \\ 
		\cline{2-14}
		 & MF & 44.1   & 75.7  &  54.8   & 90.0 &  40.7  & 73.1 &  25.5  & 50.5    &  53.2  &  85.2 & 46.2  & 79.4  \\ 
	 \hline
		
	  \multirow{2}{*}{TransT~\cite{chen2021transformer}}  &EF  &  47.4  &  74.2 & 58.8  & 84.7   &  64.4 & 95.3  &  43.9  & 70.5  & 54.7  & 84.0 & 51.8 & 79.1 \\ 
	\cline{2-14}
		  & MF  &   49.5 & 74.7   & 49.1   & 73.7 & 57.4  & 87.1   & 28.6 & 49.3  &  54.7   & 83.7 &   49.3 & 76.2  \\ 
		  \hline
		
		 
		\multirow{2}{*}{ToMP~\cite{mayer2022transforming}}  &EF &  32.0  & 50.6  & 61.8 & 89.5   &  56.3   & 79.5 &  31.1  & 47.7  &  43.0  & 60.9 & 52.0   & 76.0 \\ 

		\cline{2-14}
		 & MF  &  47.7  &  76.8 &  56.6   & 86.4 & 61.8 & 94.4   & 44.8  & 84.8  &  55.5  & 87.3 & 52.3  & 83.1 \\ 
		 \hline
		DeT~\cite{yan2021det}  & -  &  52.5  &  78.8 &  57.3   & 86.7 & 65.9 &    96.0 & 58.2  &  95.4 &  56.4  & 82.5 & 54.2  &  81.2 \\ 
		 \hline
		HMFT~\cite{zhang2022visible}  & -  &  40.2  &  67.7  & 51.4  &    86.7 & 52.6  &87.7  &  46.9  & 82.5  & 54.9   & 90.3 &  49.1 & 84.6  \\ 
		 \hline
		 FENet~\cite{zhang2021object}  & -  &  53.1  & 83.5 & 58.2   & 83.9    & 62.5& 94.7 &  47.2  & 72.4  &   57.8 & 88.5 &  55.6 & 84.3  \\ 
    
        \hline
        \textbf{AFNet} (Ours)   & -  &  \textbf{55.5}  & \textbf{84.9}  &   \textbf{64.7}  &  \textbf{93.8} &   \textbf{66.3} & \textbf{96.4} &  \textbf{62.0}  & \textbf{98.8}  &  \textbf{60.1}  &\textbf{90.3} &  \textbf{58.4} & \textbf{87.0} \\ 
		\hline
		\hline
	\end{tabular}}
	\caption{	 Attribute-based RSR/RPR scores(\%) on FE240hz~\cite{zhang2021object} dataset against state-of-the-art trackers.}
	\label{tab:overall}
	\vspace{-0.4cm}
\end{table*}

\setlength{\tabcolsep}{2.2pt}
\begin{table}[tbp]
	\centering
    \scalebox{0.95}{
	\begin{tabular}{l|c|cc|cc|cc}
		\hline
		\hline

		\multirow{2}{*}{Methods} & Fusion &
		\multicolumn{2}{c|}{ Rigid  }  & \multicolumn{2}{c|}{Non-Rigid  }  & \multicolumn{2}{c}{All } \\
		  \cline{3-8}
		  
		& Type &   RSR  & RPR  & RSR   & RPR & RSR   & RPR   \\ 
		 \hline
		\multirow{2}{*}{ATOM~\cite{danelljan2019atom}}  & EF  & 45.2  & 58.1   & 22.4  & 30.6 &  36.8   & 48.0 \\ 
		\cline{2-8}
		& MF & 47.9  & 61.1    & 20.7 & 27.8 &    37.9 &  48.9 \\ 
	   \hline
	   \multirow{2}{*}{DiMP~\cite{bhat2019learning}}  & EF   & 49.3  &63.6    & 25.4 & 36.8 &  40.5 &  53.8  \\ 
		\cline{2-8}
		& MF & 50.1 & 65.5   & 27.8 & 39.5 & 41.9   & 56.0  \\ 
	   \hline
        \multirow{2}{*}{PrDiMP~\cite{danelljan2020probabilistic}}  & EF & 46.5  &  65.3  & 31.0 & 45.8 &  40.8   & 58.2  \\ 
		\cline{2-8}
		& MF &  47.2 &    60.9 & 23.6 & 33.1  &  38.5   & 50.7  \\ 
	   \hline
	   \multirow{2}{*}{STARKs~\cite{yan2021learning}}  & EF   & 50.0   &    63.7 & 26.7  &  37.2 &    41.5 & 54.0  \\ 
		\cline{2-8}
		& MF  & 50.1 &    64.0 & 27.4  & 38.3   &  41.8   & 54.6  \\ 
	   \hline
	   \multirow{2}{*}{TransT~\cite{chen2021transformer}}  & EF   &  43.1 &   59.6  & 25.4 & 38.5 &   36.6  & 51.9  \\ 
		\cline{2-8}
		& MF  & 43.9  & 63.6   & 26.7 &  40.3  &  37.6   & 55.1  \\ 
	   \hline
	   
	   \multirow{2}{*}{ToMP~\cite{mayer2022transforming}}  & EF  & 45.2  & 57.3   & 20.2 & 27.7 &    36.0 &  46.4 \\ 
		\cline{2-8}
		& MF &  46.7 &    59.5 & 23.0 & 31.8 &   38.1  &  49.4 \\ 
		\hline
		DeT~\cite{yan2021det}  & -  &  48.9   & 62.8    & 33.3  & 45.5  &  43.2     & 56.6 \\
		\hline
		HMFT~\cite{zhang2022visible}    & -   &  50.0 & 64.0  & 27.2 & 39.7   &  41.6  &  55.1   \\ 
		\hline
		FENet~\cite{zhang2021object}   & -   &  \textbf{51.0} &    65.9 &  32.3 & 46.7  &  44.2   & 58.9  \\ 
    
        \hline
        \textbf{AFNet} (Ours)   & -   &  50.8  & \textbf{66.1}    &  \textbf{33.4} & \textbf{47.6} &  \textbf{44.5}   & \textbf{59.3}   \\ 
		\hline
		\hline
	\end{tabular}
	}
	\caption{State-of-the-art comparison  of   rigid   and non-rigid targets on the VisEvent ~\cite{wang2021viseventbenchmark} dataset.
	}
	\label{tab:vis}
	\vspace{-0.5cm}
\end{table}

\subsection{Comparison with State-of-the-art Trackers}
To demonstrate the effectiveness of our method, we compare AFNet with the nine state-of-the-art trackers. Specifically, ATOM~\cite{danelljan2019atom}, DiMP~\cite{bhat2019learning},  PrDiMP~\cite{danelljan2020probabilistic}, STARKs~\cite{yan2021learning},
TransT~\cite{chen2021transformer} and ToMP~\cite{mayer2022transforming} are conventional frame-based trackers. For a fair comparison, we extend them to multi-modality trackers via the following two fusion strategies: (i) Early Fusion (EF),
we first add the aggregated events and corresponding frame
as unified data, and then feed it into trackers; (ii) Middle Fusion (MF), we first use the backbone of these trackers to extract the frame and event features separately before feeding the sum of these features into the regressor.  We also compared  three original multi-modality methods:  DeT~\cite{yan2021det},  HMFT~\cite{zhang2022visible}, and FENet~\cite{zhang2021object} are 
frame-depth, frame-thermal, and frame-event trackers, respectively. All approaches are re-trained and tested on the FE240hz and VisEvent datasets. Following~\cite{zhang2021object}, we use RSR and RPR to evaluate all trackers. RSR and RPR focus on the overlap and center distance between the ground truth and the predicted bounding box, respectively.

Figure~\ref{fig:PRSR} (a) shows the overall evaluation results on the FE240hz~\cite{zhang2021object} dataset, which demonstrates the proposed AFNet  offers state-of-the-art high frame rate tracking performance and outperforms other compared approaches in terms of both precision and success rate. In particular, our proposed AFNet achieves an  87.0\% overall RPR and 58.4\%  RSR,  outperforming the runner-up by  2.7\%  and 2.8\%, respectively. We further validate the robustness of our AFNet under five common challenging scenarios:  high dynamic range (HDR), low-light (LL), fast motion (FM), no motion (NM), and severe background motion (SBM). Among them, the first three conditions present challenges for tracking in the conventional frame modality, while the last two scenarios provide difficulties for the event modality.
As shown in Table~\ref{tab:overall}, we can see that AFNet surpasses other approaches in all conditions. 
These results validate the effectiveness of our proposed approach on high frame rate object tracking.
The extended multi-modal methods~\cite{danelljan2019atom, bhat2019learning, danelljan2020probabilistic, yan2021learning, chen2021transformer,mayer2022transforming} lack a well-designed fusion module, preventing them from  efficiently combining the complementary information of the two domains. While original multi-modality trackers  DeT~\cite{yan2021det},  HMFT~\cite{zhang2022visible} and FENet~\cite{zhang2021object} do not address the misalignment between frame and event data at different measurement rates, causing ambiguity when locating targets.
Figure~\ref{fig:results} further qualitatively shows the effectiveness of our AFNet in different challenging conditions.

\def\wsr{1.0\linewidth}
\def\wsign{0.75\linewidth}
\def\wpr{0.5\linewidth}
\begin{figure*}[htbp]
	\centering
	\small
    \scalebox{0.95}{
	\begin{tabular}{ccccccc}
 \multicolumn{7}{c}{\includegraphics[width=\wsr]{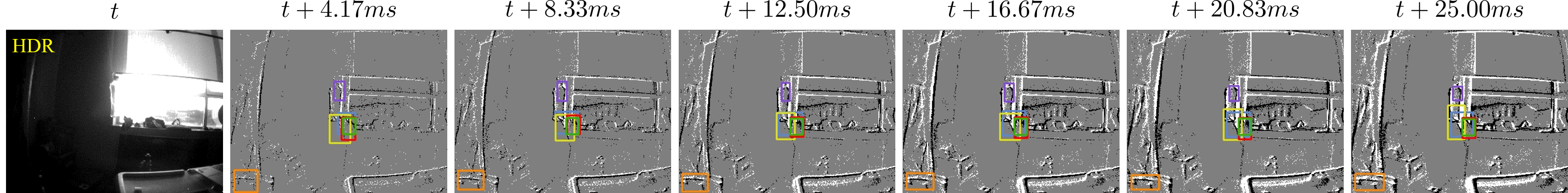}}  \\ 
 \multicolumn{7}{c}{\includegraphics[width=\wsr]{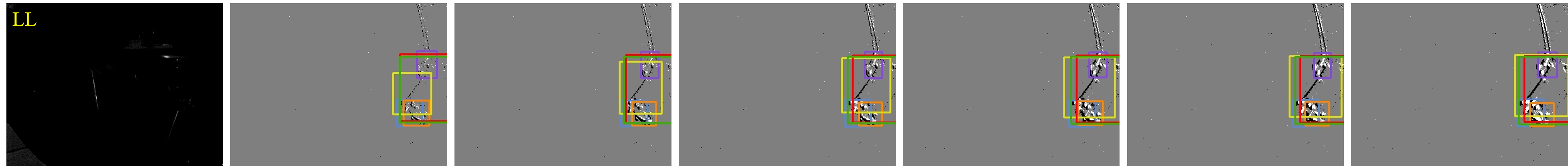}}  \\ 
 \multicolumn{7}{c}{\includegraphics[width=\wsr]{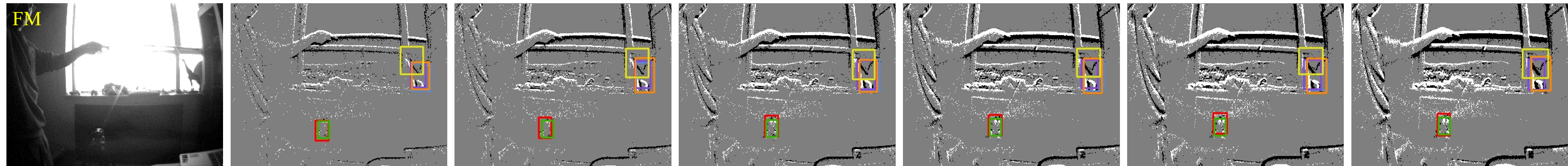}}  \\ 
 \multicolumn{7}{c}{\includegraphics[width=\wsr]{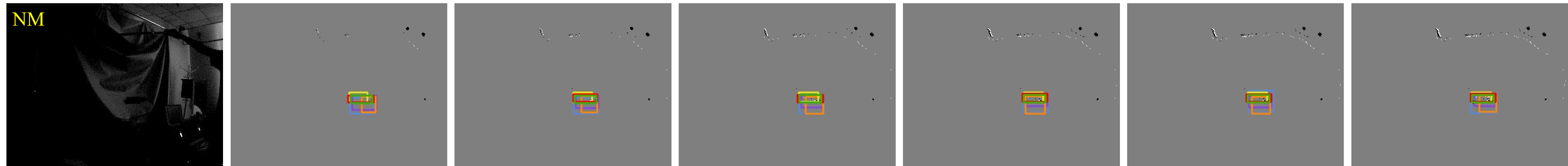}}  \\ 
 \multicolumn{7}{c}{\includegraphics[width=\wsr]{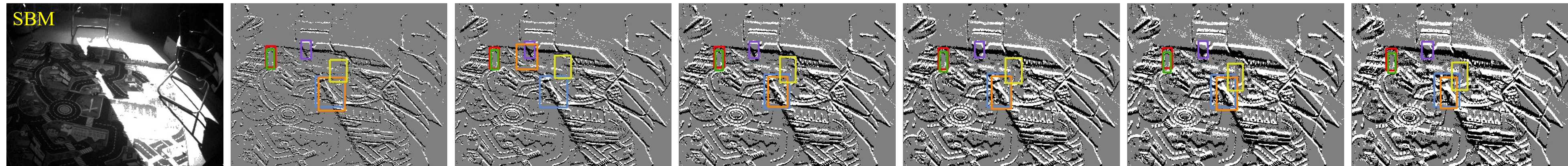}}  \\ 
 \multicolumn{7}{c}{\includegraphics[width=\wsign]{figures/visual_results/visual_sign.pdf}}  \\ 
 
 
	\end{tabular}}
    \vspace{-0.3cm}
	\caption{Qualitative comparison of AFNet against SOTA trackers on the FE240hz dataset~\cite{zhang2021object} under five challenging conditions.  All trackers locate the target  at time $t + \Delta t$ with  conventional frames at time $t$ and aggregated events  at time $t + \Delta t$ as inputs.} 
	\label{fig:results}
	\vspace{-0.4cm}
\end{figure*}

Even though the VisEvent dataset~\cite{wang2021viseventbenchmark} has a low frame rate annotation, it provides various non-rigid targets that are absent from the FE240hz dataset.  Thus, we also compare our AFNet against other state-of-the-art methods on VisEvent.  As shown in  Figure~\ref{fig:PRSR} (b), our AFNet obtains
44.5\% and 59.3\% in terms of RSR and RPR, respectively, surpassing all previous methods. Table~\ref{tab:vis}  reports the evaluation of various trackers on rigid and non-rigid targets, showing that AFNet outperforms other competing trackers on these two attributes, except the RSR on rigid targets.
These results validate that  our proposed multi-modality approach still remains effective for low frame rate frame-event tracking. 
 

\subsection{Ablation Study}

\noindent \textbf{Impact of Input Modalities.} To validate the effectiveness of fusing frame and event modalities, we design comparative experiments based only on a single  modality: (i) tracking with low frame rate conventional frames, then linearly interpolating the results to 240Hz; (ii) tracking with aggregated events of 240Hz. As shown in the rows \textit{A} and \textit{B} of Table~\ref{tab:ablation}, when using only frame or event modality as input, the performance of trackers is 16.2\%/26.9\% and 43.6\%/66.9\% at PSR/PPR, respectively.
These results are significantly worse than our AFNet, which demonstrates the necessity of multi-modality fusion for high frame rate tracking.

 \setlength{\tabcolsep}{4.0pt}
\begin{table}[tbp]
 
	\centering
    \scalebox{0.95}{
	\begin{tabular}{cl|cccc}
		\hline
		\hline
		  & Models & RSR$ \uparrow $ &OP$_{0.50}\uparrow $ &OP$_{0.75} \uparrow $  &RPR$ \uparrow $   \\
		
		\hline
	\textit{A.}	& Frame Only  & 16.2 &  15.8  &    3.4 & 26.9  \\
	\textit{B.}	& Event Only  & 43.6 & 53.4  & 18.6  &  66.9  \\
	\hline
    \textit{C.}	& w/o ECA & 55.1  & 69.3 & 29.1   &  82.4   \\
	\textit{D.}	& ECA w/o ST  &  55.8  & 69.8  & 31.2   & 83.0   \\
	\textit{E.}	& ECA w/o DA  &  55.5 & 70.0 & 30.9   & 82.8   \\
		
	\hline
	\textit{F.}	& w/o CF &  55.9  & 69.2 &  31.5  &  83.7   \\
	\textit{G.}	& CF w/o $ \mathcal{K} $ &  56.2  &  69.5 & 31.6    &  84.3   \\
	\hline
	\textit{H.}	& Ours  &\textbf{58.4}   &\textbf{73.5} &\textbf{32.6}   &\textbf{87.0}    \\
		\hline
		\hline
	\end{tabular}
	}
	\caption{Ablation study results.}
	\label{tab:ablation}
	\vspace{-0.6cm}
\end{table}

\noindent \textbf{Influence of  Event-guided Cross-modality  Alignment (ECA).}
Our proposed ECA module has two key components: style transformer (ST) and deformable alignment (DA). We thus conduct the following experiments to validate the effectiveness of ECA: (i) without ECA; Inside ECA, (ii) without ST (ECA w/o ST); (iii) without DA (ECA w/o DA). We retrain these three modified models, and the corresponding results are shown in the rows \textit{C-E} of Table~\ref{tab:ablation}, respectively. We can see that the proposed ECA module and its components all contribute to the tracking performance of AFNet. When the ST is removed, the PSR and PPR drop significantly by 2.6\% and 4.0\%, respectively. This illustrates that combining the frame  modality's content with the event modality's style plays a key role in multi-modality alignment.
The performance drops by 2.9\%/4.2\% at PSR/PPR when the DA is removed. This drop demonstrates that cross-frame-rate alignment between conventional frames and events indeed decreases target location ambiguity and enhances the discrimination ability of our tracker. 
To  further verify  cross-modality and cross-frame-rate alignment capabilities of ECA, we  visualize the feature heatmaps of the frame modality  prior to and following ECA. As shown in Figure~\ref{fig:fea_eca},  the first example shows a target that is moving upwards. We can see that the frame features shift the attention to the location of the aggregated events by our ECA.  The second illustration shows that frame features suffered from the HDR scenario. With our ECA, target  location ambiguity is eliminated. The aligned frame features will be fused with event features to 
improve the discriminative ability of our tracker further.

\def\wsr{0.95\linewidth}
\def\wpr{0.5\linewidth}
\begin{figure}[tbp]
	\centering
	\small
	\begin{tabular}{ccccc}
	\multicolumn{5}{c}{\includegraphics[width=\wsr]{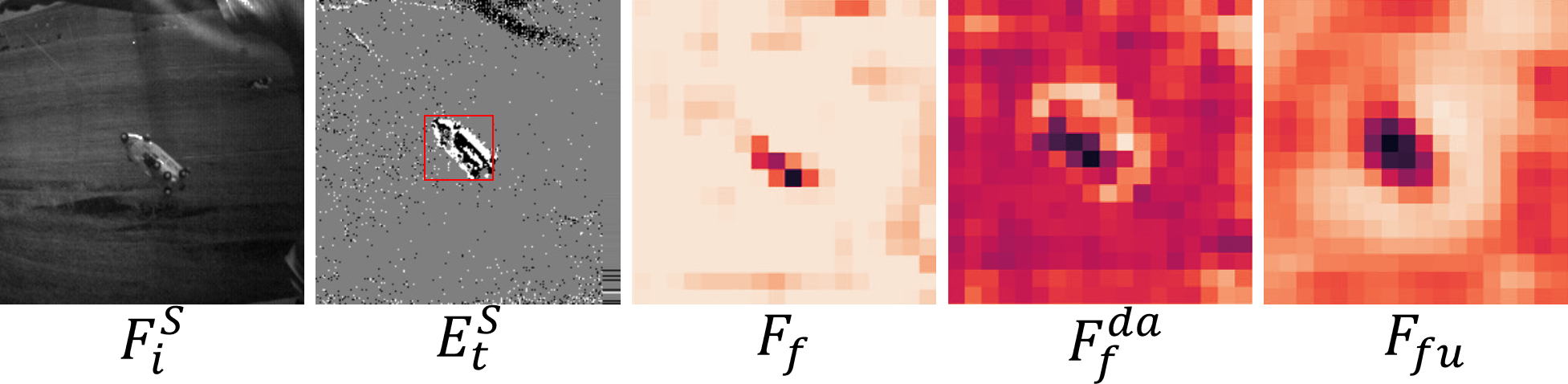}}   \\ 
	\multicolumn{5}{c}{\includegraphics[width=\wsr]{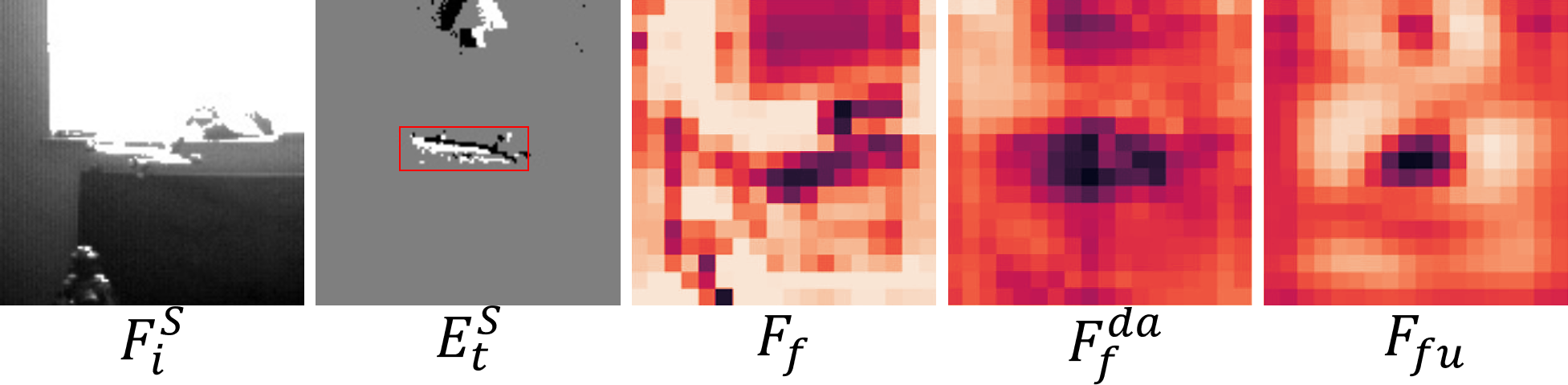}}   \\ 
 
	\end{tabular}
    \vspace{-0.3cm}
	\caption{Visualization of  features from the frame modality  before (\textit{i.e.}, $F_f$) and after (\textit{i.e.}, $F_f^{da}$) alignment by our ECA. $F_i^S$ and $E_t^S$ are the frame modality input and event modality input of the search branch, respectively. $F_{fu}$ denotes the final fused feature.} 
	\label{fig:fea_eca}
	\vspace{-0.5cm}
\end{figure}

\noindent \textbf{Influence of Cross-correlation Fusion (CF).}
We assess the influence of our CF module by replacing it with a concatenation operation in our AFNet. As shown in the row \textit{F} of Table~\ref{tab:ablation} ,  the performance drops on PSR and PPR by 2.5\% and 3.3\%  illustrate that a  well-designed multi-modality fusion strategy is essential. We further validate the impact of  cross-correlation between two modalities by removing the dynamic filter. The results in the row \textit{G} of Table~\ref{tab:ablation} demonstrate that complementing one modality with information from another indeed enhances the feature representation.

\noindent \textbf{Event Representation.}
We provide  ablation on the way events are converted to frames from two perspectives:
(i) The frame rate of accumulated event frames.  We conduct experiments with different event
frame rates on the FE240hz dataset. The results in Table~\ref{tab:framerate} indicate that AFNet
performs the best at all six event frame rates.
(ii) The starting point of accumulation. We report the performance of accumulating events since the last event frame (a) and since the last intensity frame (b), see Table~\ref{tab:accumulation}. The results of (a) on the top-3 methods are clearly lower than (b). This is because the accumulation method (a) leads to too sparse event frames, while (b) provides more motion cues for tracking.

\setlength{\tabcolsep}{3.0pt}
\begin{table}[tbp]
	\centering
	\small
    \scalebox{1.0}{
	\begin{tabular}{c|cccccc}
		\hline
      \hline
       Event Frame Rate (Hz)  &  40  & 80    & 120    & 160 & 200  &  240   \\
     \hline
      DeT~\cite{yan2021det}  & 49.7 & 52.2 &  51.3 & 53.5  & 54.3  & 54.2 \\
      FENet~\cite{zhang2021object}  & 52.4  & 54.4 & 55.6  & 52.8 & 54.7  & 55.6  \\
      \hline
      AFNeT  & \textbf{56.1} & \textbf{56.5} &  \textbf{58.0}  & \textbf{57.4} &  \textbf{57.9} & \textbf{58.4} \\
        \hline
		\hline
	\end{tabular}
	}
    \vspace{-0.3cm}
	\caption{RSR of various event frame rates on the top-3 trackers.}
	\label{tab:framerate}
	\vspace{-0.6cm}
\end{table}

\noindent \textbf{High Frame Rate Tracking Based on Interpolation.} One question in our mind is whether interpolation on  results or conventional frames still yields satisfactory high frame rate tracking results. To answer this question, we conduct two interpolation strategies: (i)  We first aggregate events at the frame rate of  conventional frames. Then, these aggregated events and frames are utilized for training and testing trackers to predict low frame rate results, which are further linearly interpolated to generate high frame rate 
bounding boxes.
(ii) We employ the video interpolation approach SuperSloMo~\cite{jiang2018super} on conventional frames to predict high frame rate sequences for evaluation. Take note that the input of the event branch of all multi-modality  trackers is replaced with interpolated frames. 
As shown in Figure~\ref{fig:inter}, the results of interpolating on low frame rate results and on conventional frames  are both  noticeably inferior to using high frame rate aggregated events. These results demonstrate that designing multi-modality alignment and fusion networks to fully exploit the high temporal resolution of events for achieving high frame rate tracking is a feasible and significant manner.

\setlength{\tabcolsep}{1.5pt}
\begin{table}[htbp]
	\centering
	\small
    \scalebox{1.0}{
	\begin{tabular}{c|ccc|ccc}
		
	\hline
    \hline
    
	& \multicolumn{3}{c|}{\textbf{(a)} since last event frame} & \multicolumn{3}{c}{\textbf{(b)} since last intensity frame} \\
	\cline{2-7}
	& DeT~\cite{yan2021det} & FENet~\cite{zhang2021object} & AFNet & DeT~\cite{yan2021det} & FENet~\cite{zhang2021object} & AFNet \\
 
    \hline
    RSR & 49.8 & 51.8  &54.9 & 54.2  & 55.6 & 58.4  \\
    RPR & 75.5 &78.4  & 83.2 & 81.2 & 84.3 & 87.0 \\
	\hline
    \hline
	\end{tabular}
	}
	\vspace{-0.2cm}
	\caption{Comparison of start times for event accumulation.}
	\label{tab:accumulation}
	\vspace{-0.5cm}
\end{table}

\subsection{Discussion}
Ideally, the tracking frame rate of our AFNet can reach the measurement rate of an event-based camera. Constrained by the existing annotated rates,  we verify the effectiveness of our proposed AFNet on FE240hz at 240Hz and VisEvent at 25Hz.   Our current focus is on exploiting multi-modality alignment and  fusion schemes for  effective and robust high frame rate tracking in various challenging conditions. However, we have not developed a lightweight network or a simple regression mechanism to speed up the evaluation of our approach.
As shown in Table~\ref{tab:speed}, we report the RPR and RSR with respect to the evaluation speed of the four multi-modality approaches on the FE240hz~\cite{zhang2021object} dataset. We can see that, at nearly equal assessment speeds, our AFNet offers the best tracking accuracy.
 
\def\wsr{0.48\linewidth}
\def\wpr{0.48\linewidth}
\def\hradar{1.2in}
\begin{figure}[tbp]
	\setlength{\tabcolsep}{1.0pt}
	\centering
	\small
	\begin{tabular}{cc}
	\includegraphics[width=\wsr]{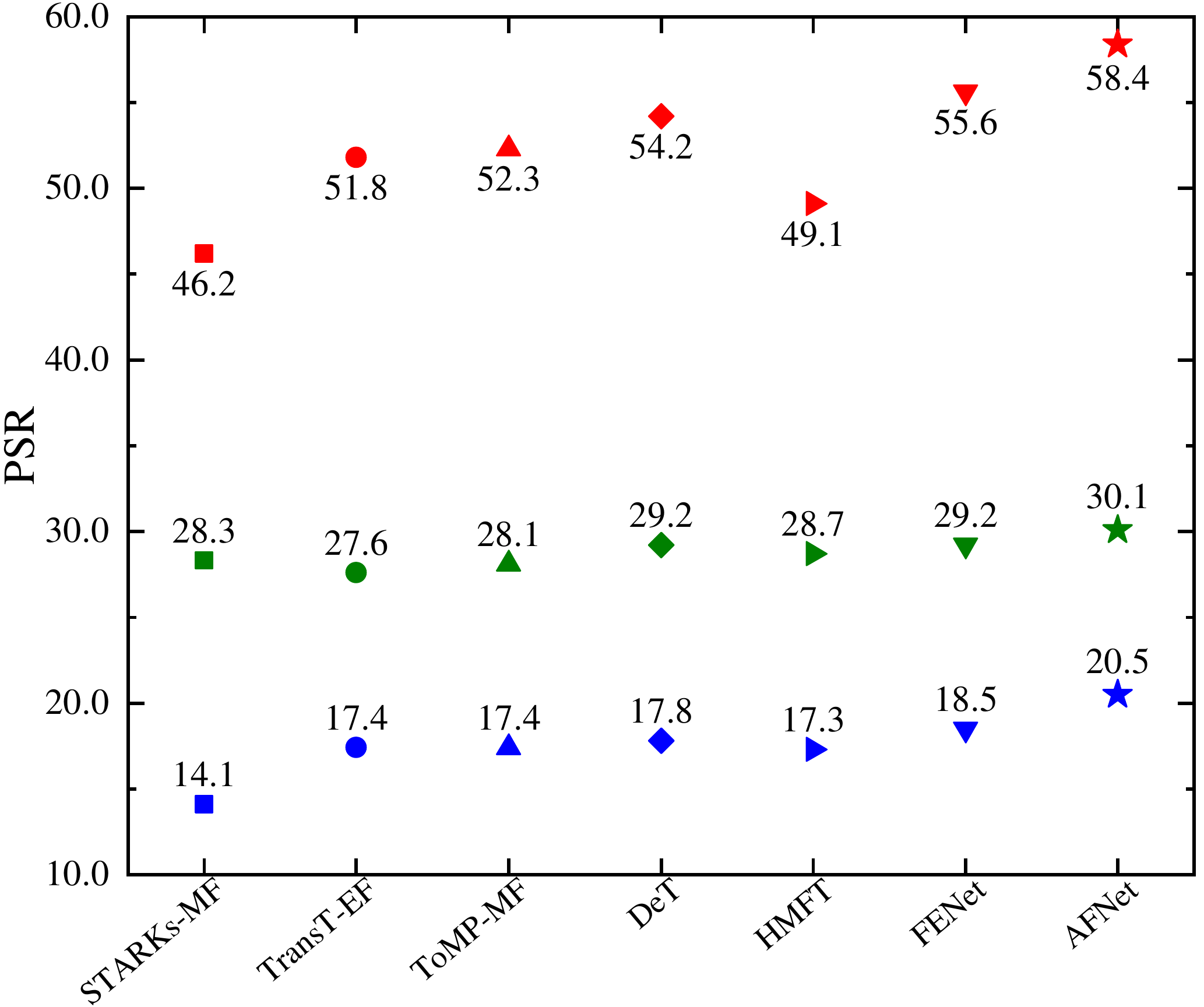}
	&
		\includegraphics[width=\wpr]{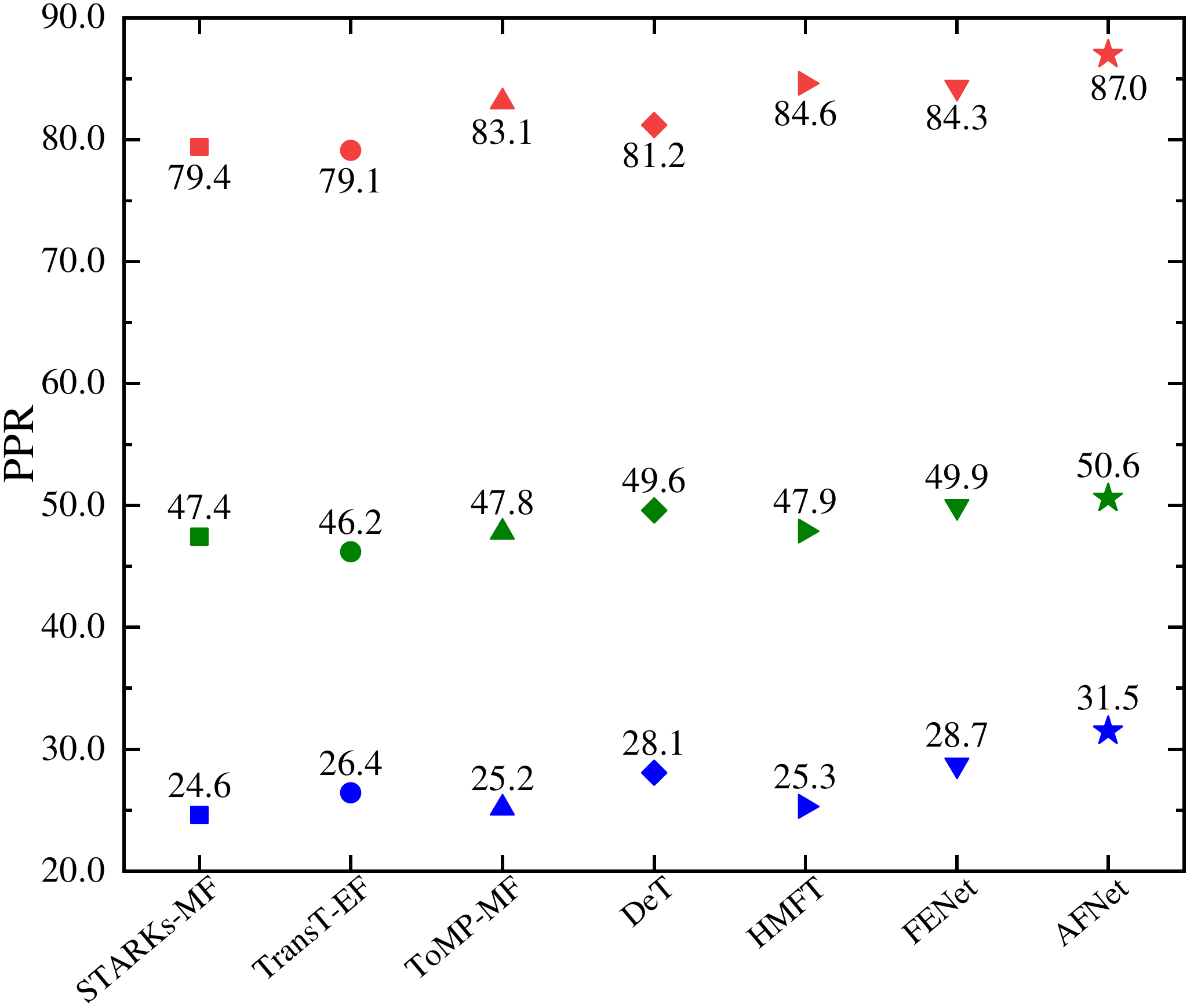}  \\ 
 
	\end{tabular}
 \vspace{-0.3cm}
	\caption{Comparison of whether to interpolate on the top-7 trackers. The blue denotes linearly interpolated performance on low frame rate tracking results; The green is tracking results on high frame rate conventional frames interpolated by SuperSloMo~\cite{jiang2018super};  While red represents the results of utilizing aggregated events that  have a higher frame rate than conventional frames.} 
	\label{fig:inter}
	\vspace{-0.4cm}
\end{figure}

\setlength{\tabcolsep}{3.0pt}
\begin{table}[htbp]
    \small
	\centering
    \scalebox{1.0}{
	\begin{tabular}{c|cccc}
		\hline
		\hline
       Methods & DeT~\cite{yan2021det} & HMFT~\cite{zhang2022visible} & FENet~\cite{zhang2021object} & AFNet \\
        \hline
        RSR & 54.2 & 49.1 & 55.6 & \textbf{58.4} \\
        RPR & 81.2 & 84.6 & 84.3 &  \textbf{87.0} \\
        Speed (FPS) & \textbf{36.68} & 34.83  &  35.5 & 36.21  \\
		\hline
		\hline
	\end{tabular}
	}
 \vspace{-0.2cm}
	\caption{Comparison of accuracy and efficiency of multi-modality approaches on the FE240hz~\cite{zhang2021object} dataset.}
	\label{tab:speed}
	\vspace{-0.45cm}
\end{table}

\section{Conclusion}

In this paper, we propose a multi-modality architecture for  high frame rate single object tracking, which  is comprised  of two key components: event-guided cross-modality alignment (ECA) module and cross-correlation fusion (CF) module. The novel-designed ECA scheme is able to effectively establish cross-modality and cross-frame-rate alignment between conventional frames and aggregated events at the  feature level. After alignment, the CF module focuses on fusing the advantages of both modalities by complementing one modality with information from another. Extensive experiments  and ablation validation demonstrate the effectiveness and
robustness of our AFNet in various challenging conditions.   The proposed AFNet
is the first in a line of work that jointly exploits frame and event modalities for high frame rate object tracking.

\noindent
\textbf{Acknowledgements.} This work was supported in part by National Key Research and Development Program of China (2022ZD0210500), the National Natural Science Foundation of China under Grant  61972067/U21A20491/U1908214, the HiSilicon(Shanghai) Technologies Co.,Ltd (No. TC20210510004), and the Distinguished Young Scholars Funding of Dalian (No. 2022RJ01).

\clearpage
{\small
\bibliographystyle{ieee_fullname}
\bibliography{main}
}

\end{document}